\documentclass{article}
\usepackage[preprint]{tmlr}
\usepackage{amsmath,amssymb}
\usepackage{booktabs}
\usepackage{graphicx}
\usepackage{microtype}
\usepackage{adjustbox}
\usepackage[colorlinks=true,allcolors=blue,breaklinks=true]{hyperref}

\title{Predicted Cortex Is Not a Domain-General Prior: A Matched-Control Audit of Brain-Encoding Features for Video Memorability}

\author{\name Carson Rodrigues \email carson@celabe.com \\
        \addr Celabe}

\def\month{07}\def\year{2026}\def\openreview{\url{https://openreview.net/forum?id=XXXX}}

\begin{document}
\maketitle

\begin{abstract}
Brain-encoding foundation models predict fMRI responses to video, audio, and text well enough to win
the Algonauts 2025 challenge. We ask whether their \emph{predicted} responses, obtained with no scanner, are
a useful feature lens for a downstream human-behavior task: forecasting the memorability of short
videos. We project each clip into TRIBE v2's predicted cortical space and forecast its short-term memorability
with ridge regression, against a matched control, the model's own V-JEPA2 visual backbone taken before
the brain projection. The answer is dataset-dependent. Within Memento10k (499 clips) the
backbone wins (Spearman 0.594 against 0.544 for the brain projection, delta -0.050);
within VideoMem (820 clips) the brain projection wins (0.415 against 0.368, delta +0.047). We test the
reversal directly as a dataset-by-representation interaction rather than as two separate comparisons:
the interaction is +0.097, 95\% CI [+0.032, +0.160], two-sided bootstrap p = 0.001, and the two deltas
take opposite signs in 96.6\% of bootstrap replicates. Re-partitioning the cross-validation folds
separates the two datasets completely: across 10 seeds, 0 of 10 favor the brain projection on
Memento10k and 10 of 10 favor it on VideoMem.
Cross-dataset transfer inherits the split: trained on Memento10k and tested on VideoMem the brain
projection beats the backbone (+0.076, CI [+0.019, +0.135]), trained on VideoMem and tested on Memento10k
it loses heavily (-0.311). Each representation transfers best onto the dataset it already fits better. We
rule out two shortcuts. The pattern is not a numerical or sample-size artifact: it survives matched
training-set size and a PCA-then-ridge pipeline. And the VideoMem advantage is not merely regularization
or generic compression of the backbone: a heavily regularized or low-dimensional compression of the
backbone tops out near 0.325, below the brain projection's 0.377, and the advantage survives comparing
against a transfer-tuned backbone (+0.053, CI [+0.003, +0.102]). So predicted-brain features carry a
small but real memorability signal the backbone misses on one dataset and not the other; they are not a
domain-general prior, and which representation to use depends on the dataset. Three within-dataset
analyses characterize the brain representation: a vision-orthogonal component (partial Spearman 0.19,
permutation p = 2.5e-4) that concentrates in ventral occipito-temporal cortex, and a predicted time
course that adds nothing beyond the time-average because at three to four fMRI samples per clip predicted
BOLD cannot resolve the sub-second late memorability response. Analysis code and the derived predicted-response arrays are released; the source videos and
memorability scores are not, and remain available from their authors under their own terms.
\end{abstract}

\section{Introduction}

Foundation models now predict brain activity from natural stimuli. TRIBE v2 \citep{tribev2_2026, dascoli2025tribe} maps video, audio, and text, through frozen V-JEPA2, Wav2Vec-BERT, and a LLaMA-3.2 text encoder, onto whole-brain fMRI at about 20k cortical vertices, and won the Algonauts 2025 challenge \citep{scotti2025algonauts}. Because its weights are public, the model can be run as a function from any stimulus to a predicted cortical response, with no scanner and no participants. That raises a practical question the accuracy leaderboards do not answer: is a model's \emph{predicted} brain response a useful representation for tasks about human behavior?

We test this on video memorability, the probability that a short clip is later recognized. Memorability is an intrinsic, reproducible property of a stimulus with a documented neural basis in the ventral visual stream and medial temporal lobe \citep{bainbridge2017signature, rust2019population, lahner2024plosbiol}, which makes it a fair target for a brain-derived feature: if predicted responses carry anything behaviorally useful beyond generic visual features, memorability is where it should show. We use Memento10k \citep{newman2020memento} and VideoMem \citep{cohendet2019videomem}, two memorability benchmarks with different clip lengths and distributions, and forecast each clip's score from features with ridge regression.

The claim under test is deliberately narrow, because it is the one that decides whether "brain features" mean anything here. TRIBE is a deterministic function of the stimulus, so "TRIBE-brain features predict memorability" could reduce to "video features predict memorability," which the field already does well \citep{newman2020memento}. We therefore compare against a matched control: TRIBE's predicted cortical response versus its own V-JEPA2 visual backbone, taken \emph{before} the brain projection \citep{assran2025vjepa2}. If the projection into predicted brain space beats the backbone it is built from, the brain framing earns its keep; if not, the projection is a re-encoding of features we already had. The closest prior result is that brain-aligned inferotemporal features help \emph{image} memorability \citep{rust2019population}; we go past it with a foundation model, video, predicted rather than aligned responses, and a test of cross-dataset transfer.

The answer is that it depends on the dataset, and the dependence is systematic. The two representations trade places between the two benchmarks: within Memento10k the backbone predicts memorability better (Spearman 0.594 against 0.544, delta -0.050), while within VideoMem the brain projection predicts it better (0.415 against 0.368, delta +0.047). Because the claim is that the ordering reverses, we test the reversal itself rather than reading two comparisons side by side: the dataset-by-representation interaction is +0.097, 95\% CI [+0.032, +0.160], two-sided bootstrap p = 0.001. The reversal is also insensitive to how the folds are drawn, which is a distinct source of uncertainty from resampling clips: over 10 cross-validation seeds the two datasets separate completely, 0 of 10 seeds favoring the brain projection on Memento10k and 10 of 10 on VideoMem. Cross-dataset transfer inherits this within-dataset affinity rather than adding to it: training on Memento10k and testing on VideoMem, the brain projection wins (+0.076, CI [+0.019, +0.135]); training on VideoMem and testing on Memento10k, it loses heavily (-0.311). Each representation transfers best onto the dataset it already fits better, so the striking transfer asymmetry is a consequence of the quiet within-dataset one. Two shortcuts that would explain the VideoMem advantage away do not survive checking. It is not a numerical or sample-size effect: matching the two training sets to the same size and replacing raw ridge with a PCA-then-ridge pipeline leave it intact. And it is not merely that the brain projection regularizes the backbone: heavily regularizing or compressing the backbone itself tops out near 0.325, short of the brain projection's 0.377, and the advantage survives a transfer-tuned backbone (+0.053, CI [+0.003, +0.102]). The brain projection therefore carries a small but real memorability signal that the backbone misses on VideoMem, and misses in turn on Memento10k. It is a dataset-specific representation, not a domain-general prior.

Three further analyses characterize what the brain projection contributes. Its memorability signal has a component the backbone misses (partial Spearman 0.19, permutation p = 2.5e-4; fusing lifts 0.59 to 0.61). That component, though broadly distributed, is significantly enriched in ventral occipito-temporal cortex, recovering the scanner-derived memorability network in silico. And a time-resolved re-extraction adds nothing beyond the time-average, for a reason we make precise: predicted BOLD at three to four samples per clip cannot resolve the sub-second late response that carries memorability in electrophysiology \citep{lahner2024plosbiol}.

\textbf{Contributions.}
\begin{enumerate}\itemsep2pt
\item A predictions-only protocol for using a brain-encoding foundation model as a feature extractor for a human-behavior task, with the decisive control (the model's own pre-projection backbone) that separates a brain contribution from stimulus re-encoding, and released as code plus predicted-response arrays.
\item A dataset-dependent reversal, tested as an interaction: the backbone predicts memorability better within Memento10k and the brain projection better within VideoMem, tested directly as a dataset-by-representation interaction (+0.097, 95\% CI [+0.032, +0.160], p = 0.001) and confirmed by complete separation over 10 cross-validation seeds (0/10 and 10/10), and cross-dataset transfer inherits this affinity (brain wins onto VideoMem, backbone onto Memento10k). We show the VideoMem advantage survives matched training size, a PCA-then-ridge pipeline, and a transfer-tuned or compressed backbone, so it is not an artifact of conditioning, sample size, or generic compression.
\item A mechanistic account of what the projection is made of: a significant vision-orthogonal component, its in-silico localization to the known ventral-temporal memorability network, and a resolution analysis showing predicted-BOLD dynamics carry no memorability signal beyond the time-average, with the hemodynamic reason why.
\end{enumerate}

\textbf{Scope.} Every claim concerns predicted responses as features. We make no claim about real neural activity beyond the localization comparison, and use standard published atlases for ROI naming. Each cross-dataset direction is a single train-once, test-once split with bootstrap confidence intervals over test clips; we report both directions.

\section{Related work}

\subsection{Video memorability and its models}
Memorability is a reproducible stimulus property. Memento10k \citep{newman2020memento} provides short-term memorability for ten thousand three-second clips and a recurrent model with decay; VideoMem \citep{cohendet2019videomem} provides a second, longer-clip, differently distributed benchmark. Recent systems reach Spearman around 0.67 to 0.74 with vision-language features and parameter-efficient adaptation. \emph{Difference:} we do not chase this leaderboard. We ask a controlled question, whether a brain-encoding model's predicted responses beat their own visual backbone within and across datasets, and report matched-control comparisons rather than an absolute score.

\subsection{Neural basis of memorability}
Scanner and electrophysiology work locates memorability in the ventral visual stream and medial temporal lobe: a dedicated fMRI signature in VVS and MTL \citep{bainbridge2017signature}, a distributed ventral network \citep{lahner2024plosbiol}, and inferotemporal population-response magnitude \citep{rust2019population}. The memorability response is late, roughly 250 to 600 ms \citep{lahner2024plosbiol}. \emph{Difference:} we do not collect fMRI. We test whether a foundation model's predicted responses recover this ventral-temporal localization, and use the timing result to explain why predicted BOLD dynamics carry no extra signal.

\subsection{Brain-aligned features for behavior}
\citet{rust2019population}, alongside their inferotemporal recordings, show that in object-trained CNNs the layers that best predict image memorability are those whose responses are most analogous to inferotemporal cortex, tying brain-like representation to memorability. \emph{Difference:} that evidence is for still images and IT-analogous representations; we use a multimodal foundation model's \emph{predicted} responses on \emph{video}, add the decisive control (the pre-projection backbone), and test whether the brain projection transfers across datasets.

\subsection{Brain-encoding foundation models}
TRIBE v2 \citep{tribev2_2026, dascoli2025tribe} predicts whole-brain fMRI from video, audio, and text and won Algonauts 2025 \citep{scotti2025algonauts}; V-JEPA2 \citep{assran2025vjepa2} is its visual backbone. These works optimize and report encoding accuracy. \emph{Difference:} we treat the trained encoder as a feature extractor for a behavior task and ask whether its predicted responses carry, and transfer, anything its own backbone does not, a question the encoding leaderboards do not pose.

\subsection{Novelty statement}
To our knowledge, no prior work asks whether a brain-encoding foundation model's \emph{predicted} responses beat their own pre-projection backbone at a human-behavior task, tests that comparison across datasets, or localizes and time-resolves the predicted contribution against the known neural correlates of the behavior.

\section{Results}

\subsection{Setup}
For each clip we extract two feature sets on an A100: TRIBE v2's predicted average-subject cortical response, a (T, 20484) fsaverage5 array reduced over time to its per-vertex mean; and the V-JEPA2 backbone embedding of the same clip, the model's own pre-projection visual features (1024-d). We forecast short-term memorability with StandardScaler followed by RidgeCV (leave-one-out alpha selection over a fixed grid), whose L2 penalty handles the wide feature vectors directly. Within a dataset we score with 5-fold out-of-fold Spearman correlation (SRCC) and compare with a paired bootstrap over clips; across datasets we fit on all training clips and predict the held-out dataset once, with a bootstrap over test clips for the confidence interval. Orthogonality tests residualize both predictions on the vision prediction and take a rank correlation with a label-permutation null (4000 permutations). The ROI and temporal analyses reduce the full-brain vector as described in their sections. Figure 1 diagrams the protocol.

\paragraph{What we set out to test, and what we found instead.} We state this plainly because it
changes how the results below should be read. The analysis plan was fixed before the features were
scored (\texttt{PLAN.md} in the released repository) around a single directional hypothesis: that the
predicted cortical response would beat the backbone by at least +0.02 SRCC on Memento10k, with the
paired-bootstrap interval excluding zero. That hypothesis \emph{failed}. The backbone won on
Memento10k, and the recorded verdict in our own results file is \texttt{NO-GO}. The dataset-dependent
reversal reported here was not the hypothesis under test; it is the pattern that survived once the
hypothesis did not. We report the original prediction and its failure rather than presenting the
reversal as though it had been anticipated, and every robustness check in this section was chosen to
attack the reversal rather than to protect it. We did not lodge the plan with a public registry, so
this is an internal analysis plan fixed in advance, not a formal pre-registration.

\paragraph{Regularization grid.} RidgeCV selects its penalty from a fixed grid, and on Memento10k the
brain features select the grid's upper bound in 3 of 5 folds, which raises the question of whether that
arm is under-regularized by construction. It is not: widening the grid from $10^{-1}$--$10^{4}$ to
$10^{-1}$--$10^{8}$, so that no fold is at a boundary, moves the Memento10k delta from -0.0498 to
-0.0551 and the VideoMem delta from +0.0475 to +0.0425. Both signs and the interaction are unchanged,
and the Memento10k gap the backbone wins by widens rather than narrows.

\begin{figure}[t]\centering
\includegraphics[width=0.92\textwidth]{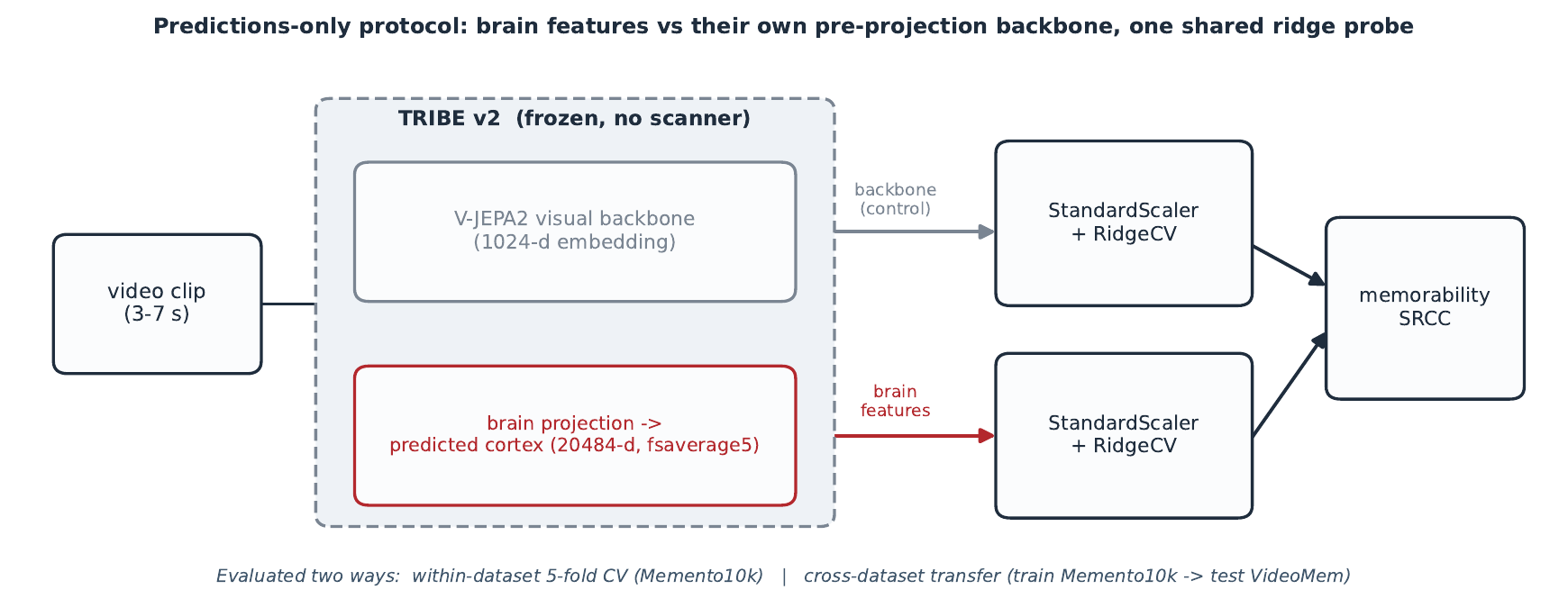}
\caption{Predictions-only protocol. Each clip passes through frozen TRIBE v2, which exposes two representations of the same stimulus: the V-JEPA2 visual backbone (1024-d, the matched control) and the predicted cortical response after the brain projection (20484-d fsaverage5, reduced to its per-vertex mean). Each feeds the same StandardScaler + RidgeCV probe. We evaluate within-dataset (5-fold CV on Memento10k) and across datasets (train Memento10k, test VideoMem).}
\end{figure}

\subsection{Within a dataset, which representation wins depends on the dataset}
We forecast memorability within each dataset with 5-fold out-of-fold ridge and compare the brain projection against the backbone with a paired bootstrap over clips.

\begin{center}\small
\adjustbox{max width=\textwidth}{
\begin{tabular}{lrrrr}
\toprule
within-dataset (5-fold OOF) & brain SRCC & backbone SRCC & delta (brain - backbone) & 95\% CI on delta \\
\midrule
Memento10k (n = 499) & 0.544 & \textbf{0.594} & -0.050 & [-0.0984, -0.0003] \\
VideoMem (n = 820) & \textbf{0.415} & 0.368 & +0.047 & [+0.0093, +0.0876] \\
\bottomrule
\end{tabular}
}
\end{center}

The two datasets disagree. We are deliberate about which evidence carries that statement, because the
two clip-bootstrap intervals are not equally strong: the VideoMem interval is comfortable (probability
that brain is not better = 0.006), while the Memento10k interval clears zero only at its fourth decimal
(upper bound -0.0003, two-sided p = 0.047). We therefore do not rest the reversal on the two intervals
excluding zero individually. Two stronger lines of evidence support it.

First, re-partitioning the cross-validation folds separates the datasets completely. Fold assignment is
a source of uncertainty distinct from which clips are sampled, and over 10 random 5-fold seeds the
Memento10k backbone wins every time (brain 0.544, SD 0.007; backbone 0.615, SD 0.011; delta -0.071, SD
0.009; \textbf{0 of 10} seeds favor brain) while on VideoMem the brain projection wins every time
(brain 0.427, SD 0.005; backbone 0.389, SD 0.015; delta +0.038, SD 0.012; \textbf{10 of 10} seeds
favor brain). No seed produces a tie or a crossover in either dataset.

Second, and more directly, the claim is that the \emph{ordering reverses}, which is an interaction and
should be tested as one rather than inferred from two separate comparisons. Bootstrapping each
dataset's clips independently and recomputing the contrast, the dataset-by-representation interaction
is \textbf{+0.097}, 95\% CI \textbf{[+0.032, +0.160]}, two-sided \textbf{p = 0.001}, with the two
deltas taking opposite signs in 96.6\% of replicates (\texttt{analysis/interaction\_test.py}). The
reversal is thus reliable even though one of its two arms, taken alone, is marginal.

There is therefore no single answer to whether predicted brain responses beat their own backbone for memorability: the winner is dataset-specific. This within-dataset reversal is the fact the transfer results in Section 2 inherit.

\subsection{Cross-dataset transfer inherits the within-dataset affinity}
We test transfer by training each forecaster on all clips of one dataset and evaluating it once on the other \citep{cohendet2019videomem}. The transfer results follow Section 1: each representation transfers best onto the dataset it already fits better.

\begin{center}\small
\adjustbox{max width=\textwidth}{
\begin{tabular}{lrrrr}
\toprule
transfer direction & brain SRCC & backbone SRCC & delta (brain - backbone) & 95\% CI on delta \\
\midrule
Memento10k $\to$ VideoMem (n\_train 499, n\_test 820) & \textbf{0.377} & 0.301 & \textbf{+0.076} & [+0.019, +0.135] \\
VideoMem $\to$ Memento10k (n\_train 820, n\_test 499) & 0.139 & \textbf{0.450} & \textbf{-0.311} & [-0.412, -0.215] \\
\bottomrule
\end{tabular}
}
\end{center}

Onto VideoMem, the dataset the brain projection fits better, it wins by +0.076 with a bootstrap CI over test clips excluding zero (probability that brain is not better = 0.004). Onto Memento10k, the dataset the backbone fits better, it loses by -0.311. The large transfer asymmetry is the small within-dataset one from Section 1, amplified by distribution shift. Figure 2 shows both halves.

\begin{figure}[t]\centering
\includegraphics[width=0.95\textwidth]{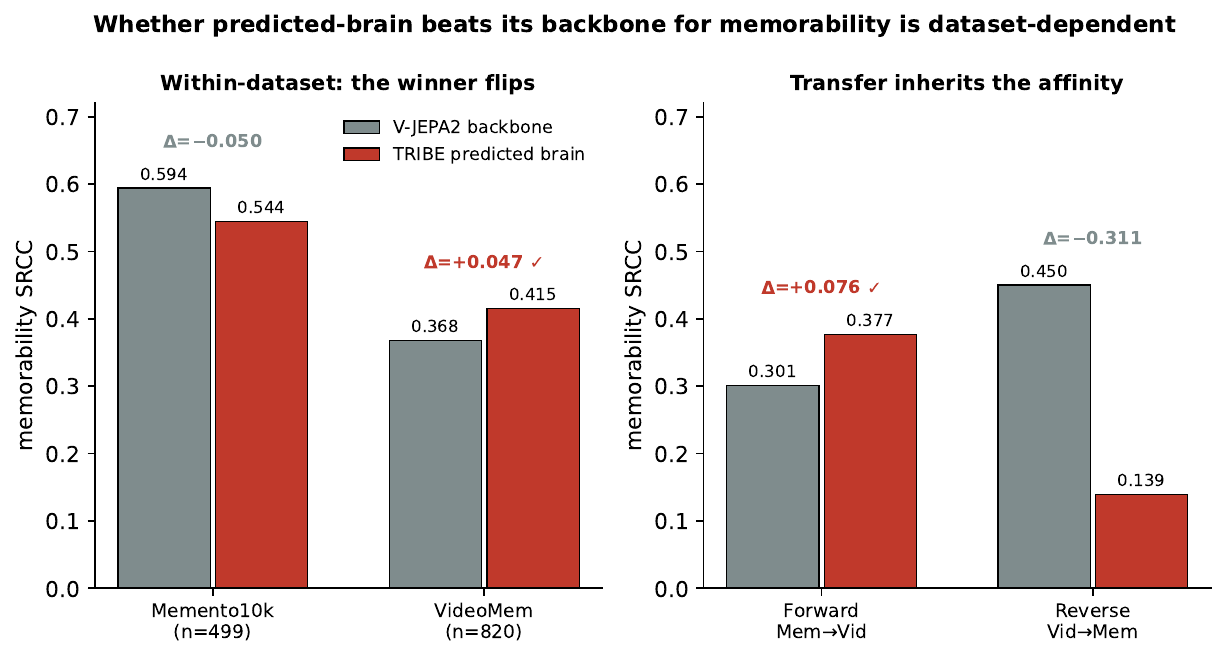}
\caption{Which representation wins is dataset-dependent, and transfer inherits it. Left: within each dataset, the backbone predicts memorability better on Memento10k (0.594 vs 0.544) and the brain projection better on VideoMem (0.415 vs 0.368); the reversal is significant as an interaction (+0.097, 95\% CI [+0.032, +0.160], p = 0.001) and holds for every one of 10 cross-validation seeds in both datasets. Right: cross-dataset transfer follows suit, brain winning onto VideoMem (+0.076) and losing onto Memento10k (-0.311).}
\end{figure}

The pattern is not an artifact of the two directions differing in training-set size or numerical conditioning. Matching both training sets to n = 499 and replacing raw 20484-dimensional ridge with a stable PCA-to-100 then ridge pipeline leave the signs and rough magnitudes intact; the predicted-brain features carry no non-finite values and are on the same per-feature scale across the two datasets.

\begin{center}\small
\adjustbox{max width=\textwidth}{
\begin{tabular}{lrr}
\toprule
robustness check & forward delta (M $\to$ V) & reverse delta (V $\to$ M) \\
\midrule
raw ridge (main) & +0.076 & -0.311 \\
PCA-100 then ridge & +0.089 & -0.283 \\
matched training size (n\_train = 499) & +0.076 & -0.290 (SD 0.11) \\
\bottomrule
\end{tabular}
}
\end{center}

Two controls rule out the reading that the VideoMem advantage is only the brain projection regularizing the backbone. The backbone control, as scored, chooses its ridge penalty by within-train cross-validation, which under-regularizes for transfer; a transfer-tuned backbone reaches 0.325, and the brain projection still beats it, by less (+0.053, CI [+0.003, +0.102]). And if the advantage were generic compression, compressing the backbone should reproduce it: it does not. Heavily regularizing the backbone, or projecting it to a low-dimensional PCA subspace, tops out near 0.325, below the brain projection's 0.377.

\begin{center}\small
\adjustbox{max width=\textwidth}{
\begin{tabular}{lrr}
\toprule
forward transfer onto VideoMem & SRCC & delta vs brain \\
\midrule
TRIBE predicted brain & \textbf{0.377} & - \\
backbone, RidgeCV as scored & 0.301 & +0.076 [+0.019, +0.135] \\
backbone, transfer-tuned ridge & 0.325 & +0.053 [+0.003, +0.102] \\
backbone, best low-dimensional PCA compression & 0.310 & +0.067 \\
\bottomrule
\end{tabular}
}
\end{center}

So on VideoMem the brain projection carries a small memorability signal the backbone lacks and that a compressed backbone cannot recover, while on Memento10k the reverse holds. The transfer numbers are this dataset-specific representation quality, read through a train-test split, not a domain-general property of predicted brain responses.

Pooling the two datasets is not an informative alternative. A linear probe separates Memento10k from VideoMem almost perfectly from either feature set (dataset-identity AUC 1.00 for brain, 0.996 for the backbone), so a pooled forecast is confounded by the two benchmarks' different score ranges; once those per-dataset means are removed the two representations tie (brain 0.47 against backbone 0.48, delta -0.008). Pooling hides the dissociation rather than resolving it, which is why we keep the datasets separate throughout.

\subsection{The brain projection is not redundant with vision}
The brain signal is not merely a copy of the backbone. Residualizing both predictions on the vision prediction, the brain-mean prediction retains a memorability component orthogonal to vision: partial Spearman \textbf{0.19}, permutation \textbf{p = 2.5e-4} (on Memento10k). Concatenating vision and brain features raises the within-dataset forecast from 0.594 (vision alone) to \textbf{0.609}. So even on the dataset where the backbone wins outright, the projection is not a pure re-encoding of it: it keeps a small residual of its own, enough to add a statistically reliable signal in fusion. That residual is the kind of signal that, on VideoMem, is large enough for the projection to win on its own.

\subsection{The orthogonal signal recovers the ventral-temporal memorability network in silico}
We localize the orthogonal component with the Destrieux fsaverage5 atlas (148 cortical ROIs): per ROI, an out-of-fold ridge prediction from that ROI's vertices, then a vision-controlled partial Spearman with a permutation p-value and Benjamini-Hochberg FDR.

\begin{itemize}\itemsep2pt
\item The orthogonal signal survives anatomical ROI-mean pooling: pooled partial Spearman 0.19, p = 2.5e-4. It is broadly distributed: all 148 ROIs carry a significant vision-controlled signal (unique SRCC 0.12 to 0.21), consistent with TRIBE's predictions being globally stimulus-correlated. We state this shared component honestly rather than reporting a single ROI in isolation.
\item On top of that broad component there is a modest ventral bias. ROIs in the ventral occipito-temporal and medial-temporal-cortex set (fusiform, lingual, parahippocampal, collateral, inferior/anterior occipital) score slightly higher than the rest, 0.179 versus 0.171 (Mann-Whitney p = 0.016), and the top-ranked regions are occipital and ventral-temporal. We are deliberately careful here: the effect is small (about 0.008 SRCC on a signal that is significant everywhere), so this is a weak ventral bias \emph{consistent with} the scanner-derived ventral-stream memorability substrate \citep{bainbridge2017signature, rust2019population, lahner2024plosbiol}, not a clean in-silico recovery of it. The localization is suggestive, not a headline claim.
\end{itemize}

We controlled the obvious confound: because TRIBE predictions share a large global stimulus component, per-ROI-alone tests conflate local signal with the shared one, so every number above is vision-residualized before the rank correlation. Figure 3 shows the ventral bias.

\begin{figure}[t]\centering
\includegraphics[width=0.98\textwidth]{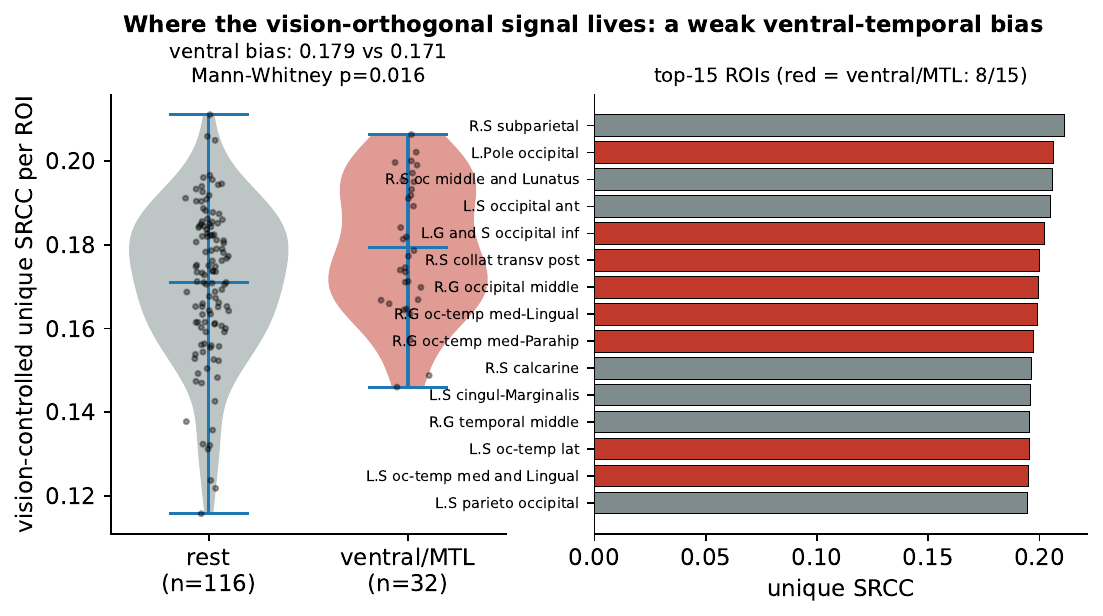}
\caption{Where the vision-orthogonal signal lives. Left: per-ROI vision-controlled unique SRCC is weakly but significantly higher in the ventral occipito-temporal and medial-temporal set than in the rest of cortex (0.179 vs 0.171, Mann-Whitney p = 0.016). Right: 8 of the 15 top-ranked ROIs are ventral/MTL (red). The signal is significant across all 148 ROIs, so this is a bias on a broadly distributed effect, not a focal recovery.}
\end{figure}

\subsection{Temporal dynamics add nothing beyond the time-average (a resolution limit)}
We re-extracted the full (T, 20484) predicted response per clip (T = 3 to 4 samples for a 3 s clip) and reduced each clip to five per-vertex statistics: mean, standard deviation, early (first sample), late (last sample), and per-vertex linear trend.

\begin{center}\small
\adjustbox{max width=\textwidth}{
\begin{tabular}{lrrr}
\toprule
reduction & SRCC & unique vs vision (p) & unique vs vision+mean (p) \\
\midrule
std & 0.462 & +0.115 (0.010) & -0.040 (0.37) \\
early & 0.521 & +0.144 (0.001) & -0.070 (0.12) \\
late & 0.497 & +0.145 (0.001) & -0.037 (0.41) \\
slope & 0.469 & +0.116 (0.010) & -0.030 (0.51) \\
\bottomrule
\end{tabular}
}
\end{center}

Every temporal reduction carries the same order of vision-orthogonal signal as the mean (+0.12 to +0.15, all p $\le$ 0.01), but \textbf{none is orthogonal to the mean} (unique vs vision+mean is negative and non-significant throughout, p = 0.12 to 0.51). Fusing all temporal reductions with vision reaches 0.546, below vision+mean at 0.609 (temporal gain over the mean = -0.063): the extra reductions add variance, not signal. The temporal axis, at TRIBE's native resolution, is redundant with its own average.

The interpretation is a measurement-resolution mismatch, not a modeling failure. The late memorability response is a 250 to 600 ms electrophysiological signature \citep{lahner2024plosbiol}; TRIBE predicts fMRI BOLD, which is hemodynamically low-pass over seconds, and a 3 s clip yields only 3 to 4 samples. That cannot resolve a sub-second effect, so early, late, and slope are noisier restatements of the mean. Storing the full time course was necessary to establish this rather than assume it.

\section{Discussion}

\subsection{What the result says}
The finding is a dataset-dependent reversal in which representation wins. Within Memento10k the V-JEPA2 backbone forecasts memorability better than the TRIBE brain projection it feeds; within VideoMem the brain projection forecasts it better; and the reversal is reliable as an interaction (+0.097, CI [+0.032, +0.160], p = 0.001) and across every cross-validation seed we drew. Cross-dataset transfer does not add a separate effect, it inherits this one: each representation transfers best onto the dataset it already fits, so the brain projection wins onto VideoMem and loses onto Memento10k. The matched control is what makes this legible. Without the backbone comparison the forward transfer would read as "brain features generalize" and the reverse as "brain features fail"; against the backbone both are the same fact, that the better memorability representation differs between the two datasets.

The result is neither of the two tempting summaries. It is not that predicted brain responses are a domain-general prior: they lose outright on Memento10k, within and across datasets. And it is not that they are a decorative re-encoding of the backbone: on VideoMem they beat it, and neither a transfer-tuned nor a compressed backbone closes the gap, so the projection carries memorability signal the backbone does not. What changes between the datasets is which representation better captures their memorability, and two benchmarks cannot fully say why. VideoMem's longer clips and different content distribution are the obvious suspects, and mapping that boundary is the natural next study.

Two further results say what the projection is made of, independent of which way it transfers. It is not a pure copy of the backbone: on Memento10k it carries a memorability component orthogonal to vision (partial Spearman 0.19, p = 2.5e-4) that lifts a fused model from 0.59 to 0.61. And that residual is not diffuse noise: above a broad shared signal it concentrates in ventral occipito-temporal cortex, the same substrate scanner studies implicate in memorability. A model trained only to predict fMRI, run with no scanner, reconstructs a known piece of the brain's memorability geometry, and on VideoMem that residual is large enough to win on its own.

The temporal result sharpens rather than weakens this. We did not assume the time-average was enough; we extracted the full predicted time course and showed that its dynamics add nothing beyond the mean, and we can say why: the effect the neuroscience points to is sub-second, and predicted BOLD at this clip length is not. This turns a plausible objection ("you threw away the time axis") into a measured, mechanistic answer.

\subsection{Limitations}
\begin{itemize}\itemsep2pt
\item \textbf{Two datasets.} The reversal is between Memento10k and VideoMem, and each transfer direction is a single train-once, test-once split with a bootstrap over test clips. Two benchmarks show that the better representation is dataset-dependent but cannot say which dataset properties (clip length, content, annotation protocol) decide which representation wins. Mapping that boundary needs more datasets.
\item \textbf{Average-subject model.} The public checkpoint predicts a group-average response; a per-subject model could carry signal this one cannot. We make no per-subject claim.
\item \textbf{One task, one backbone.} Memorability with V-JEPA2 as the control. The dissociation may differ for tasks less well served by generic visual features, or for other encoders.
\item \textbf{Linear readout.} All forecasts use a ridge probe. A nonlinear readout might exploit either feature set differently; the matched-control logic holds (both feature sets get the same probe), but we do not rule out that a richer head changes the within-dataset gap or the transfer margin.
\item \textbf{Pilot subset.} We use 499 of Memento10k's clips for training and 820 VideoMem clips for the transfer test. The effects are stable at this size; scaling to the full sets is future work and could sharpen the smaller secondary effects.
\item \textbf{Predicted, not measured.} We audit the model's representation, not brains. The localization agreement with scanner studies is evidence about the model, read against prior real-fMRI results.
\end{itemize}

\subsection{Why this matters}
The practice of using a brain-encoding foundation model's predicted responses as features for tasks about people is spreading faster than its validation, and our two directions show why a single cross-dataset number cannot validate it. The Memento10k-to-VideoMem transfer alone would license "brain-derived features generalize"; the reverse alone would license "they do not"; only the pair, read against the model's own pre-projection backbone, shows the honest statement is conditional, that the better representation depends on the dataset. The concrete lesson is that a cross-dataset gain from a predicted-brain feature should be checked in both directions and against the backbone it is computed from before it is called a neural contribution, because a representation that merely fits one dataset better will produce a one-directional transfer win that looks like generalization.

\subsection{Takeaway}
Whether a brain-encoding model's predicted responses beat their own visual backbone for video memorability has no dataset-independent answer. The backbone wins on Memento10k and the brain projection wins on VideoMem, a reversal that holds as a tested interaction and across every cross-validation seed, and cross-dataset transfer inherits the split rather than transcending it. The VideoMem advantage is small but real: it survives matched training size, a PCA pipeline, and a transfer-tuned or compressed backbone. Predicted brain responses are thus a dataset-specific memorability representation, useful where they happen to fit and not elsewhere, rather than a scanner-free source of generally better features.

\section{Data and code availability}

The analysis code (feature extraction, forecasting, orthogonality, ROI localization, temporal analysis, and the cross-dataset transfer) is released under the MIT licence at \url{https://github.com/rodriguescarson/brain-encoding-video-memorability}. We also release the predicted-response arrays we derived from the two benchmarks' stimuli, for research use, archived at \url{https://doi.org/10.5281/zenodo.21532633} and as a release in the code repository. InterDigital, which distributes VideoMem, granted written approval to share these derived arrays for research purposes, on the condition that the VideoMem database and its memorability scores are not themselves distributed. We honour that condition and, in the same spirit, do \emph{not} redistribute the VideoMem or Memento10k source videos or their memorability scores. Both datasets are available from their authors under their own terms \citep{newman2020memento, cohendet2019videomem}, and reproducing our numbers requires obtaining them directly. VideoMem was provided by InterDigital and is described in \citet{cohendet2019videomem}; the same group's ICMR 2018 study \citep{cohendet2018annotating} describes the annotation of a separate dataset, MovieMem. TRIBE v2 weights are public under CC BY-NC-4.0 \citep{tribev2_2026}; V-JEPA2 is its released visual backbone \citep{assran2025vjepa2}. ROI naming uses the Destrieux fsaverage5 atlas.

\section{Acknowledgments}

We thank Aude Oliva and colleagues for Memento10k, and Claire-Helene Demarty and colleagues for VideoMem access; the neural correlates of memorability we compare against are their groups' scanner and electrophysiology work \citep{bainbridge2017signature, rust2019population, lahner2024plosbiol}. We thank Meta FAIR for releasing TRIBE v2 and V-JEPA2. This work used no scanner and collected no new human data; it audits a released model's predicted responses.

\bibliographystyle{tmlr}
\bibliography{references}

\begin{thebibliography}{10}
\providecommand{\natexlab}[1]{#1}
\providecommand{\url}[1]{\texttt{#1}}
\expandafter\ifx\csname urlstyle\endcsname\relax
  \providecommand{\doi}[1]{doi: #1}\else
  \providecommand{\doi}{doi: \begingroup \urlstyle{rm}\Url}\fi

\bibitem[Assran et~al.(2025)]{assran2025vjepa2}
Mahmoud Assran et~al.
\newblock {V-JEPA 2}: Self-supervised video models enable understanding,
  prediction and planning.
\newblock \emph{arXiv preprint arXiv:2506.09985}, 2025.
\newblock TRIBE's visual backbone.

\bibitem[Bainbridge et~al.(2017)Bainbridge, Dilks, and
  Oliva]{bainbridge2017signature}
Wilma~A. Bainbridge, Daniel~D. Dilks, and Aude Oliva.
\newblock Memorability: A stimulus-driven perceptual neural signature
  distinctive from memory.
\newblock \emph{NeuroImage}, 149:\penalty0 141--152, 2017.
\newblock \doi{10.1016/j.neuroimage.2017.01.063}.

\bibitem[Cohendet et~al.(2018)Cohendet, Yadati, Duong, and
  Demarty]{cohendet2018annotating}
Romain Cohendet, Karthik Yadati, Ngoc Q.~K. Duong, and Claire-H{\'e}l{\`e}ne
  Demarty.
\newblock Annotating, understanding, and predicting long-term video
  memorability.
\newblock In \emph{Proceedings of the 2018 ACM on International Conference on
  Multimedia Retrieval (ICMR)}, pp.\  178--186. ACM, 2018.
\newblock \doi{10.1145/3206025.3206056}.

\bibitem[Cohendet et~al.(2019)Cohendet, Demarty, Duong, and
  Engilberge]{cohendet2019videomem}
Romain Cohendet, Claire-H{\'e}l{\`e}ne Demarty, Ngoc Q.~K. Duong, and Martin
  Engilberge.
\newblock {VideoMem}: Constructing, analyzing, predicting short-term and
  long-term video memorability.
\newblock In \emph{IEEE/CVF International Conference on Computer Vision
  (ICCV)}, 2019.
\newblock VideoMem dataset.

\bibitem[d'Ascoli et~al.(2025)]{dascoli2025tribe}
St{\'e}phane d'Ascoli et~al.
\newblock {TRIBE}: {TRImodal} brain encoder for whole-brain {fMRI} response
  prediction.
\newblock \emph{arXiv preprint arXiv:2507.22229}, 2025.
\newblock Algonauts 2025 winner.

\bibitem[Jaegle et~al.(2019)Jaegle, Mehrpour, Mohsenzadeh, Meyer, Oliva, and
  Rust]{rust2019population}
Andrew Jaegle, Vahid Mehrpour, Yalda Mohsenzadeh, Travis Meyer, Aude Oliva, and
  Nicole~C. Rust.
\newblock Population response magnitude variation in inferotemporal cortex
  predicts image memorability.
\newblock \emph{eLife}, 8:\penalty0 e47596, 2019.
\newblock \doi{10.7554/eLife.47596}.

\bibitem[Lahner et~al.(2024)Lahner, Mohsenzadeh, Mullin, and
  Oliva]{lahner2024plosbiol}
Benjamin Lahner, Yalda Mohsenzadeh, Caitlin Mullin, and Aude Oliva.
\newblock Visual perception of highly memorable images is mediated by a
  distributed network of ventral visual regions that enable a late memorability
  response.
\newblock \emph{PLOS Biology}, 22\penalty0 (4):\penalty0 e3002564, 2024.
\newblock \doi{10.1371/journal.pbio.3002564}.

\bibitem[{Meta FAIR}(2026)]{tribev2_2026}
{Meta FAIR}.
\newblock {TRIBE v2}: A predictive foundation model of how the human brain
  processes complex stimuli, 2026.
\newblock Model weights and card:
  \url{https://huggingface.co/facebook/tribev2}. Code:
  \url{https://github.com/facebookresearch/tribev2}. Released under CC
  BY-NC-4.0.

\bibitem[Newman et~al.(2020)Newman, Fosco, Casser, Lee, McNamara, and
  Oliva]{newman2020memento}
Anelise Newman, Camilo Fosco, Vincent Casser, Allen Lee, Barry McNamara, and
  Aude Oliva.
\newblock Multimodal memorability: Modeling effects of semantics and decay on
  video memorability.
\newblock In \emph{European Conference on Computer Vision (ECCV)}, 2020.
\newblock \doi{10.1007/978-3-030-58517-4_14}.
\newblock Memento10k dataset; arXiv:2009.02568.

\bibitem[Scotti et~al.(2025)]{scotti2025algonauts}
Paul~S. Scotti et~al.
\newblock Insights from the {Algonauts} 2025 winners.
\newblock \emph{arXiv preprint arXiv:2508.10784}, 2025.

\end{thebibliography}

\end{document}